\documentclass{article}

\usepackage{nips_2017,times}

\usepackage[utf8]{inputenc} % allow utf-8 input
\usepackage[T1]{fontenc}    % use 8-bit T1 fonts
\usepackage{hyperref}       % hyperlinks
\usepackage{url}            % simple URL typesetting
\usepackage{booktabs}       % professional-quality tables
\usepackage{amsfonts}       % blackboard math symbols
\usepackage{nicefrac}       % compact symbols for 1/2, etc.
\usepackage{microtype}      % microtypography
\usepackage{booktabs} % For formal tables
\usepackage{multirow}
\usepackage{amsmath}
\usepackage{graphicx}
\usepackage[ruled]{algorithm2e} % For algorithms

\usepackage{color}
\definecolor{applegreen}{rgb}{0.55, 0.71, 0.0}

\title{SESA: Supervised Explicit Semantic Analysis}

\author{
   Dasha Bogdanova and Majid Yazdani\\
   LinkedIn Corporation \\
   Dublin, Ireland \\
  \texttt{dbogdanova,myazdani@linkedin.com}
}

\begin{document}
\nipsfinalcopy 

\maketitle

\begin{abstract}
In recent years supervised representation learning has provided state of the art or close to the state of the art results in semantic analysis tasks including ranking and information retrieval. The core idea is to learn how to embed items into a latent space such that they optimize a supervised objective in that latent space.
The dimensions of the latent space have no clear semantics, and this reduces the interpretability of the system. For example, in personalization models, it is hard to explain why a particular item is ranked high for a given user profile.
We propose a novel model of representation learning called Supervised Explicit Semantic Analysis (SESA) that is trained in a supervised fashion to embed items to a set of dimensions with explicit semantics. The model learns to compare two objects by representing them in this explicit space, where each dimension corresponds to a concept from a knowledge base. This work extends Explicit Semantic Analysis (ESA)~\cite{gabrilovich2007computing} with a supervised model for ranking problems. 
We apply this model to the task of Job-Profile relevance in LinkedIn in which a set of skills defines our explicit dimensions of the space. Every profile and job are encoded to this set of skills their similarity is calculated in this space.  We use RNNs to embed text input into this space. 
In addition to interpretability, our model makes use of the web-scale collaborative skills data that is provided by users for each LinkedIn profile. Our model provides state of the art result while it remains interpretable.
\end{abstract}

\section{Introduction and related work}
%Most methods for obtaining such vector representations rely on the \textbf{distributional hypothesis}, that states that words that occur in similar contexts tend to have similar meanings \citep{harris1954}.
%Early methods represented words as one-hot vectors aka bag-of-words, i.e. vectors with zeros at all dimensions except for the one corresponding to that word. In order to obtain a representation for a document, the vectors of all the words it contained were summed and sometimes normalised. For instance, the tf-idf model~\citep{salton:1986} normalises the word frequencies by their frequencies in the whole corpus, in order to diminish the weights of most common words. %\TODO{more examples of count-based methods: LSA? LDA?}
%These models produced very high-dimensional though sparse representations. 
%These representations suffer from several limitations. First, they do not account for similarity between different words. It is logical to expect that the words \textit{cat} and  \textit{kitten} should be more similar to each other, than the words \textit{cat} and \textit{car}, however, the one-hot representation imposes all words to be equally similar to each other.
%Second, using extremely high-dimensional representations in machine learning leads to the curse of dimensionality.

%% unsupervised methods
Feature selection is one of the most cumbersome tasks in creating a machine learning system. Representation learning tries to automate this task by creating numerical vectors that best reflect the semantics of the objects for a given task. These vectors can then be fed to machine learning algorithms. 
%In the area of natural language processing (NLP), bag-of-words has been a straightforward way of representing documents.
Most methods for obtaining vector representations of words and documents rely on the distributional hypothesis which states that words in similar contexts have similar meanings~\cite{harris1954}. Early methods were mostly count-based, i.e. relied on term frequencies in different contexts as the representation of meaning. One of the earliest and most known examples is the TF-IDF vector space model~\cite{salton:1986} that normalizes the term frequencies by inverted document frequencies to reduce the weights of the terms that frequently appear in all documents. The main shortcoming of the TF-IDF model is the assumption of independence among dimensions (words); words have different types and degrees of relationships among each other and the independence assumption is too simplistic for this case.
Latent space methods were proposed to address this issue. For example, another count-based technique for inferring vector representations is Latent Semantic Analysis (LSA)~\cite{deerwester1990indexing}, that also starts with a high-dimensional term-frequency matrix and maps it to a latent low-dimensional space by applying Singular Value Decomposition (SVD). Other methods include a probabilistic version of latent semantic analysis~\cite{hofmann1999probabilistic} and Latent Dirichlet Allocation~\cite{blei2003latent}.
More recently, the predictive methods for modeling semantics have gained popularity. These methods treat the task of mapping a word to a meaningful vector as a predictive machine learning task instead of relying on word counts. For instance, the Continuous Bag of Words and the Skip-gram models~\cite{mikolov2013efficient} of the widely used \texttt{word2vec} tool.  These predictive methods have been shown to beat the count-based ones in most Natural Language Processing (NLP) tasks~\cite{baroni2014don}. The \texttt{word2vec} models were extended to learn document representations
\cite{leandmikolov:2014}. In contrast to words in TF-IDF model, the dimensions of these latent space models have no clear meaning, resulting sometimes in hard to interpret results and difficult to introspect machine learned systems.  Interpretability of the machine learning methods has become an issue, as many applications concern not only about the prediction being correct but also about the reasons that decision has been made~\cite{li2016visualizing}.

%% unsupervised - interpretable: ESA
Explicit Semantic Analysis (ESA)~\cite{gabrilovich2007computing} tries to address this issue. It represents words as vectors in which each dimension corresponds to a knowledge base entity that is usually a Wikipedia article. It builds an inverted index of word frequencies in Wikipedia pages; each word is represented as a vector of the size of Wikipedia articles, such that the weight of each dimension is the word frequency in the corresponding Wikipedia article. To get a representation of a document, one can average the representations of all the words in that document. 

%% supervised works : PSI , DSSM ,..
All the above representation learning methods are unsupervised, and while providing us with a generic representation of the objects, they usually need to be optimized for a specific task.  In recent years many supervised representation learning models were proposed for relevance, ranking and information retrieval tasks. Polynomial Semantic Indexing (PSI) can be viewed as a supervised version of LSA that is trained for ranking~\cite{PSI}. Similarly, in \cite{WestonImageWords} images and words are embedded to a same latent space for image tagging task. Deep Semantic Similarity Model (DSSM) has been used in information retrieval and web search ranking, and also ad selection/relevance, contextual entity search, and interestingness tasks ~\cite{DSSM, DSSM-conv,shen2014latent}. These supervised representation learning models provide state of the art for ranking and relevance tasks, but remain uninterpretable. We propose an interpretable alternative for supervised representation learning for ranking and relevance tasks by extending ESA algorithm.

\begin{table}
\centering
\begin{tabular}{ p{4cm}|p{3cm}|p{4cm} }
\hline
\multicolumn{3}{ c }{Representation Learning Algorithms} \\
\hline
Supervised/Unsupervised & Interpretable & Uninterpretable \\ \hline
\multirow{2}{*}{Unsupervised} & TF-IDF~\cite{salton:1986} & LSA~\cite{deerwester1990indexing}, PLSA~\cite{hofmann1999probabilistic} \\

  & ESA~\cite{gabrilovich2007computing} & 
  LDA~\cite{blei2003latent}, \texttt{word2vec}~\cite{mikolov2013efficient}  \\ \hline
  
\multirow{2}{*}{Supervised} & \bf{SESA} & DSSM ~\cite{DSSM, DSSM-conv,shen2014latent} \\
 &  & PSI~\cite{PSI} \\ \hline
\end{tabular}
\caption{Representation learning algorithms}
\label{RLalgs}
\end{table}

Table~\ref{RLalgs} categorizes the representation learning algorithms regarding supervision signal they use and their interpretability. The rest of this paper describes SESA more in detail and presents ongoing experiments on the job-profile relevance task.

\section{SESA: Supervised Explicit Semantic Analysis}

SESA represents objects in a space 
%consisting of a set of explicit categories in which 
where each dimension has a human interpretable semantics. The model consists of the following components:
(1) \textbf{encoder}, that maps an object to a latent space; (2) \textbf{knowledge base,} that provides the explicit categories; (3) \textbf{projector}, that projects the latent representations to the explicit space; (4) \textbf{similarity scorer}, that estimates the similarity between objects in the explicit space. To train parameters of our model we also need labeled data and a loss function. 
%Figure~\ref{fig:job_rel_sesa} shows various components of our model. 
We describe the components more in detail in the remainder of this section.

% \begin{description}
% \item[Encoder:] The encoder maps an object to a \textbf{latent} space. This could be any function but in this work we are particularly interested in Deep Neural Networks for instance recurrent neural network (RNN) in case of a text object or a Convolutional Neural Network (CNN) in case of an image. Also, an encoder can be stacked together to create a new encoder.
% \item[Knowledge Base:] The knowledge base provides the explicit categories that will serve as dimensions of the explicit semantic space. This can be a simple list of semantic categories rather than a knowledge base in a strict sense. For example, the set of skills entered by users in Linkedin forms a knowledge base that profiles and jobs can be described in that space.
% \item[Projector:] The projector projects the latent representation resulting from encoder into the \textbf{explicit} semantic space. This again can be any function, but in this work we assume simple projectors as we put the burden more on the encoder.
% \item[Similarity Scorer:] The similarity scorer estimates the similarity between the object in the explicit semantic space. In this work we assume simple similarity functions but one could think of also trainable similarity functions.\footnote{The scorer can also take into account the latent representations and even external features but this is orthogonal to the contribution of this paper.} 
% \end{description}

\subsection{Encoder}
The encoder maps an object to a \textbf{latent} space. This could be any function, but in this work, we are particularly interested in neural encoders.
% for instance recurrent neural network (RNN) in case of a text object or a Convolutional Neural Network (CNN) in case of an image. Also, an encoder can be stacked together to create a new encoder.
%An encoder represents the input as a fixed-length vector in a latent, i.e. non-interpretable, vector space. 
A typical encoder for text is a recurrent neural network (RNN), such as Long Short Term Memory network~\cite{hochreiter1997lstm} or Gated Recurrent Network~\cite{cho-EtAl:2014:EMNLP2014} that have been widely used as encoders in various areas including machine translation~\cite{cho-EtAl:2014:EMNLP2014,bahdanau2015neural,firat-cho-bengio:2016} and sentiment analysis~\cite{tang2015document}. Also, encoders can be stacked to create a new encoder.

\subsection{Knowledge Base}
A knowledge base provides the explicit categories that will serve as dimensions of the explicit semantic space. This can be a simple list of semantic categories rather than a knowledge base in a strict sense. 
% For example, 
%A knowledge base is required to represent the explicit categories that serve as the dimensions of the interpretable semantic space.
ESA uses Wikipedia as a knowledge base; every document is represented as a vector where each dimension represents the strength of association with a particular Wikipedia entity. In this paper we use the set of skills entered by users in Linkedin as the knowledge base; User profiles and jobs can be described in this space.
%will present experiments on job-candidate relevance task where we use a taxonomy of skills as a knowledge base.
\subsection{Projector}
The projector projects the resulting latent representation into the \textbf{explicit} semantic space.  We can use a simple linear projector to map the latent vector into the explicit semantic space.

Let's assume $x$ shows an object's features, the implicit representation of $x$ is given by the encoder:
\begin{equation}
\pmb{e}_{impl} = f_{enc}(\pmb{x})
\end{equation}
and the projector maps the implicit representation to the explicit space: 
\begin{equation}
\pmb{e}_{expl} = f_{proj}(\pmb{e}_{impl})
\end{equation}
The simplest projector is a linear projector:
\begin{equation}
\pmb{e}_{expl} = \pmb{Wx}
\label{eq:linproj}
\end{equation}
where $\pmb{W}$ is a $m\times n$ weight matrix, where $m$ is the dimension of the encoder's output and $n$ is the number of explicit categories.

\subsection{Similarity Scorer}
The similarity scorer estimates the similarity between objects in the explicit semantic space.\footnote{The scorer can also take into account the latent representations and even external features, but this is orthogonal to the contribution of this paper.}  As the \textit{burden of learning} lies on the encoder and the projector, there is no need in complicated similarity scoring, but one could also think of trainable similarity functions.
We suggest using a dot product or a cosine similarity.

%\begin{equation}
%cos(\pmb x, \pmb y) = \frac {\pmb x \cdot \pmb y}{||\pmb x|| \cdot ||\pmb y||}
%\end{equation}

\section{SESA for Job Relevance Task}
We evaluate SESA on the task of predicting job relevance. Given a LinkedIn profile and a LinkedIn job description, the task is to predict if the person is interested in applying for this job. As a knowledge base, we use the LinkedIn's skills. This set consists of skills that were entered by LinkedIn users for each profile. The intuition behind using SESA for the task of job-profile relevance is that every job has requirements, most of which could be expressed in terms of skills. For example,  

"\textit{We are looking for talented Data Engineers with a strong programming skills and knowledge of neural networks capable of rapid application development within an Agile environment}" would likely require 

\textit{software engineering}, \textit{machine learning}, \textit{deep learning} and \textit{Agile methodologies} and would not require \textit{budget control} or \textit{online merchandising}. 

\begin{figure*}[t]
  \includegraphics[width=\textwidth]{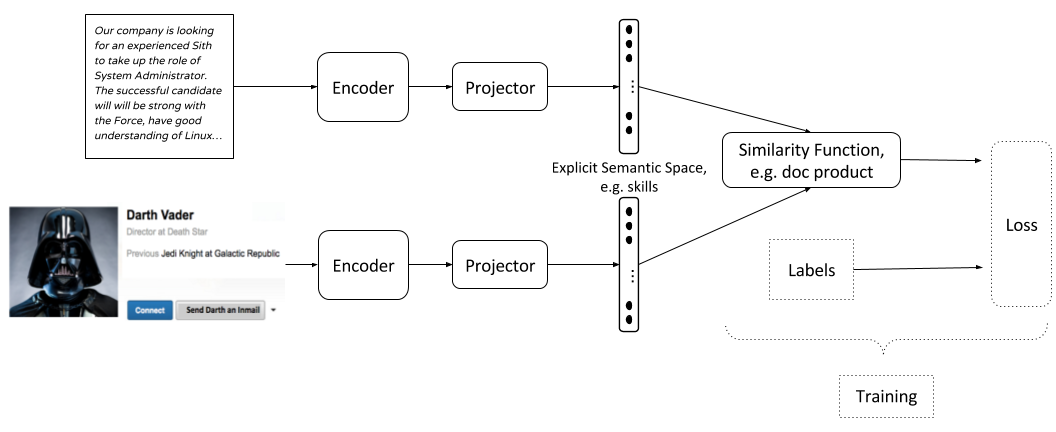}
  \caption{Illustration of SESA for job relevance. The profiles are represented in the explicit semantic space where each dimension represents a skill. The job descriptions are encoded with an LSTM and then projected to the explicit semantic space. A dot product between the explicit representations is used to predict relevance.}
  \label{fig:job_rel_sesa}
\end{figure*}

A person possessing most or all of these skills is likely to be a good match for this position, and vice versa, a person that does not possess the skills required for a job, is not likely to be a good match.
%Example skills include \textit{machine learning}, \textit{sales}, \textit{natural language processing}.
Our approach is illustrated in Figure~\ref{fig:job_rel_sesa}. We use an RNN encoder to map the job description to its latent representation and then project it to the explicit skills space. As the members' profiles are already tagged with skills, we just take those skills from the profiles. In other words, the encoder and projector in the member's side only extract the skills from the profile. Then we estimate the similarity between the explicit representations of the job and the profile.

\section{Experimental Setup}

We use a dataset of 36 million job-profile pairs, the pairs are labeled as positive if the person has applied for a job, and as negative if the person has viewed the job but did not apply. There are only 270K positive examples in the dataset.
65\% of the dataset is used for training, 5\% for validation and 30\% for testing. 
We only consider the skills that appear at least one thousand time in the training set; the final set of skills contains 35K different skills. 
%Some skills have aliases, i.e. synonym expressions referring to the same skill. 
 %Example skills include \textit{machine learning}, \textit{information retrieval}, \textit{software development}, \textit{sales}.
 %\subsection{Baselines}
We use logistic regression and gradient boosting baselines using common meta-information as the features. The features can be divided into three categories: 
(1) \textbf{job-only features}: location, company size, seniority and required skills;
(2) \textbf{profile-only features}: location, current company, current industry, gender, seniority;
(3) \textbf{job-profile similarity features}: cosine similarity and Jaccard similarity between tf-idf representation of all job information and all profile information.

There are 182 different features in total.

%\subsection{Encoder}
%We use a deterministic identity encoder for the member profile, as LinkedIn profiles include the information about one's skills.

We assume that a job description can be mapped to a set of skills required for this job. To do this, we use an LSTM encoder.
As we use the LSTM encoder, we can either use the last output vector of the RNN as the latent representation, or we can average all outputs. Our preliminary experiments have shown that averaging the outputs provides better results on the validation set. Therefore, we use this encoding strategy in the experiments we report. We use MSE loss function to train our model.

\subsection{Hyperparameters and Training}

We use an LSTM encoder with 100 units. The word embeddings are pre-trained using the skip-gram model~\cite{mikolov2013efficient} on Wikipedia and LinkedIn news and job descriptions; the dimensionality is set to 200.
The network is trained with stochastic gradient descent by minimizing mean squared error on the training set.
We use L2 regularization with regularization rate of $10^{-7}$. The batch size is set to 1000. We use early stopping on the validation set: the model is evaluated on the validation set every 500 iterations and the training is stopped if there is no improvement on the validation set for 20 consecutive iterations. The model is implemented with tensorflow.\footnote{\url{https://www.tensorflow.org/}}
Logistic regression baseline is implemented using Photon Machine Learning framework.\footnote{\url{https://github.com/linkedin/photon-ml}} It was trained for 100 iterations with regularization rate of 0.1. Gradient boosting baseline is implemented with XGBoost library,\footnote{\url{http://xgboost.readthedocs.io/}} the hyperparameters are tuned on the validation set. We report the results with the maximal depth of 5,  the learning rate of 0.1 and $\gamma$ of 0.1.

\section{Results}
We use the area under the curve (AUC) of the receiver operating characteristic (ROC) curve as the evaluation metric.
Table~\ref{tab:results} compares the performance of the SESA model with the baseline systems. We test the models that perform the best on the validation set. We compare the performance of SESA when using randomly initialized word embeddings versus the pretrained embeddings.
SESA with pretrained word embeddings achieves good results outperforming most baselines and performing at the level of gradient boosting while (1) avoiding feature engineering; (2) being interpretable and (3) providing re-usable by-products that we describe in the following section.

\begin{small}
\begin{table}[t]
\centering
\begin{tabular}{p{7cm}p{1cm}}
\toprule
\textbf{Model}  &\textbf{AUC}   \\\midrule
%Random	& Last & 0.81 \\
%Pretrained &	Last &	0.82 \\
%Random	& Average	& 0.82 \\
%Pretrained & Average & \textbf{0.86} \\\midrule
SESA (Random word embeddings)	& 0.82 \\
SESA (Pretrained word embeddings) &  \textbf{0.86} \\\midrule

%\multicolumn{2}{c}{\textbf{Baselines}} \\\midrule
{Logistic Regression} & 0.78 \\
{Gradient Boosting (500 trees)} & {0.85} \\
{Gradient Boosting (1000 trees)} & \textbf{0.86} \\\bottomrule
\end{tabular}
\caption{The AUC of the SESA with an LSTM encoder and a linear projector versus the baseline feature-based systems.
}
\label{tab:results}
\end{table}
\end{small}

\subsection{SESA by-products}

Training the SESA for job-profile relevance task provides two main by-products: (1) skills embeddings; and (2) job2skill tagger.

\paragraph{\textbf{Skills Embeddings}} The matrix \pmb{W} (see Equation \ref{eq:linproj}) can be viewed as the matrix of skills embeddings. In these embeddings, the skills which behave similarly in the job-profile matching task are closer vectors. This gives us different embeddings than other unsupervised word embeddings since the embeddings are optimized for this supervised task. Further qualitative analysis of the differences is an ongoing work.

\paragraph{\textbf{Job2skill Tagger}}
The second by-product is a job2skill tagger that tags a job description with skills. This job2skill tagger needs to be tailored further to be used as a standalone tagger. 
The output of job2skill is a real vector in the space of skills in which if a skill is irrelevant can have a negative score. The training set is highly unbalanced (more than 99\% are negative examples) and therefore the model mostly learns negative correlation of the skills for jobs. 
The negative skills are useful for the relevance prediction task, but are not the main interest of the job2skill tagger.
Also, some skills are not frequent enough in the positive pairs of the dataset, and the model can not learn much about them. 
However, this trained model can be used as an initialization and be tuned further to create a standalone job2skill tagger. This tagger has the advantage of exploiting a large scale skills and click data in comparison to a tagger trained on a handcrafted skill tagging dataset. 

Table~\ref{tab:job2skill} presents an example of the job2skill output for the job description of a software engineering intern. While the SESA by-product inferred many positive skills correctly, it also inferred several non-relevant frequent skills, i.e. \textit{treasury management} and \textit{financial services}. 
%The negative skills, i.e. the dimensions in the explicit space where the job description had negative values

%\begin{footnotesize}
\begin{table}
\centering
\begin{tabular}{p{125mm}}
\toprule
\textbf{Job Title:} Software Engineer Internship \\\midrule
\textbf{Job Description:}
The ideal candidate will be excited for the challenge to transform and think critically on many computer science disciplines including product design, usability, building APIs and user-centric online applications, business logic, scaling performance, and 24x7 reliability (...) \\\midrule
\textbf{Positive Skills:} \texttt{\color{applegreen}python}, \texttt{\color{applegreen}c},  \texttt{\color{applegreen}programming}, \texttt{\color{applegreen}Amazon RDS}, \texttt{\color{applegreen}IOS development}, \texttt{\color{red}treasury management}, \texttt{\color{red}financial services}\\\midrule
\textbf{Negative Skills:} \texttt{counterintelligence}, \texttt{e-commerce consulting},  \texttt{yoga},  \texttt{scuba diving} \\\bottomrule
\end{tabular}

\caption{Example positive and negative skills inferred by the SESA model, i.e. with highly positive and highly negative association scores in the explicit space.}
\label{tab:job2skill}
\end{table}
%\end{footnotesize}
\section{Conclusions and Future Work}
We presented SESA, a novel model of representation learning. This model is trained in a supervised manner to embed objects into an explicit space with interpretable dimensions. We presented ongoing experiments on job-profile relevance with SESA where we represent both the job and the profile in a semantic space where each dimension corresponds to a particular skill in Linkedin. 
In this case, the model also has two reusable by-products: skills embeddings and skills tagger for jobs, their effectiveness should be experimented in the downstream tasks. Besides, the model made use of a large-scale collaborative skill dataset entered by LinkedIn users.

In the future we plan to experiment and improve SESA by-products. Also, we plan to evaluate SESA on other ranking and relevance tasks plus considering various other ranking loss functions that are studied in the literature.  Finally, it is straightforward to extend the model with implicit representations in addition to the explicit ones to capture semantics beyond the explicit categories, which will make the model more robust to incomplete knowledge bases and noise.

\bibliographystyle{plain}

\end{document}